# HYBRID DEEP REINFORCEMENT LEARNING AND PLANNING FOR SAFE AND COMFORTABLE AUTOMATED DRIVING

**Dikshant Gupta**
Saarland University, Computer Science Department
Saarbruecken, Germany  dikshant2210@gmail.com

**Matthias Klusch**
German Research Center for Artificial Intelligence
Saarbruecken, Germany  matthias.klusch@dfki.de

## ABSTRACT

We present a novel hybrid learning method, HyLEAR, for solving the collision-free navigation problem for self-driving cars in POMDPs. HyLEAR leverages interposed learning to embed knowledge of a hybrid planner into a deep reinforcement learner to faster determine safe and comfortable driving policies. In particular, the hybrid planner combines pedestrian path prediction and risk-aware path planning with driving-behavior rule-based reasoning such that the driving policies also take into account, whenever possible, the ride comfort and a given set of driving-behavior rules. Our experimental performance analysis over the CARLA-CTS1 benchmark of critical traffic scenarios revealed that HyLEAR can significantly outperform the selected baselines in terms of safety and ride comfort.

## 1  Introduction

**Problem.** We consider the basic problem of collision-free navigation (CFN) of a self-driving car as, in short, to navigate on a driveable path to a given goal in minimal time and collisions with objects such as other cars or pedestrians in a partially observable traffic environment. The CFN problem can be modelled as a partially observable Markov decision process (POMDP) to be solved by the car online and subject to the given car and pedestrian model. In this work, we adopt the POMDP definition and models for the CFN problem from [17] and address the additional challenge to compute driving policies that are not only safe but, whenever possible, also passenger comfortable.

**Related work.** Current CFN methods for self-driving cars can leverage either a deep reinforcement learner (DRL) [10, 6], or an approximate POMDP planner (APPL) [15, 1], or a hybrid combination of thereof [17]. While a hybrid system of DRL-supported online planning for CFN can outperform its individual components for CFN in terms of safety [17], it may suffer from long training and online planning time. However, it is not known, whether some alternative hybrid system of planning-supported DRL for CFN can be more efficient and safe. The rule-interposed learning framework in [21] was shown to enable faster training of vanilla DRL but has not been applied to hybrid DRL for CFN yet. On the other hand, many works and user studies investigated the effect of various road and load disturbance factors including human-perceived risk [18, 9, 11] on passenger or ride comfort [19, 3, 7]. In this work, we measure ride comfort based on the load disturbance factors smoothness and human- perceived risk of the ride by the car. However, the potential of hybrid CFN methods taking ride comfort into account without compromising safety in critical scenarios remains unclear. Besides, self-driving cars are expected to navigate whenever possible in compliance with a given set of default driving-behavior or traffic rules, such as not to drive on sidewalks, or to keep the lane. In some critical situations, however, safe driving may require the violation of certain rules as experienced human drivers might do without making it a habit and still being able to explain their decision for the exceptional trajectory. In [5], a hierarchical rulebook is used for explainable (production rule) reasoning for filtering rule-compliant safe trajectories but without integrated hybrid CFN learning and ride comfort.

**Contributions.** To this end, we developed HyLEAR, a novel hybrid planning-assisted DRL method for the purpose of safe navigation of self-driving cars in POMDPs considering, whenever possible, risk-aware ride comfort and driving-behavior rule compliance. For our experiments, we created the CARLA-CTS benchmark of critical traffic scenarios based on the GIDAS accident study [2] and simulated in the driving simulator CARLA.

## 2 Method HyLEAR

**Overview.** HyLEAR consists of a soft actor-critic deep reinforcement learner, named NavSAC, that is assisted by a hybrid planner. The hybrid planner leverages functional modules for (a) risk-sensitive planning of alternative safe and short paths, (b) driving behavior rule-based reasoning for the selection of one path with minimal risk and rule violations, and (c) online POMDP planning of an optimal speed action for the next step on this path. In particular, at each time step of driving simulation during training and testing of HyLEAR in CARLA, the risk-aware path planner uses an anytime weighted hybrid A* k-path planner and a pedestrian intention estimator M2P3 [16] to determine three alternative safe and short paths together with their estimated human-perceived risk values. These risk values are computed based on the driver risk fields for the paths as in [11] together with the respective cost-maps for planning. While the first path is planned with given problem related cost-map, the planning of the other two relies on modified cost-maps with lower costs of driving on free sidewalks, respectively, additional predicted pedestrian positions. The rule-based reasoner performs hierarchical rule reasoning on a given rulebook [5] of initially four priority-ordered default driving-behavior rules (avoid driving on sidewalks, minimize risk (blockage), keep the lane, take shortest path) to select the safe path with minimal human-perceived risk and rule violations. For the next full car control action, the steering angle is extracted from this selected path and the optimal speed action is determined during training by the POMDP planner IS-DESPOT [15, 1] and during testing by the trained learner NavSAC of HyLEAR.

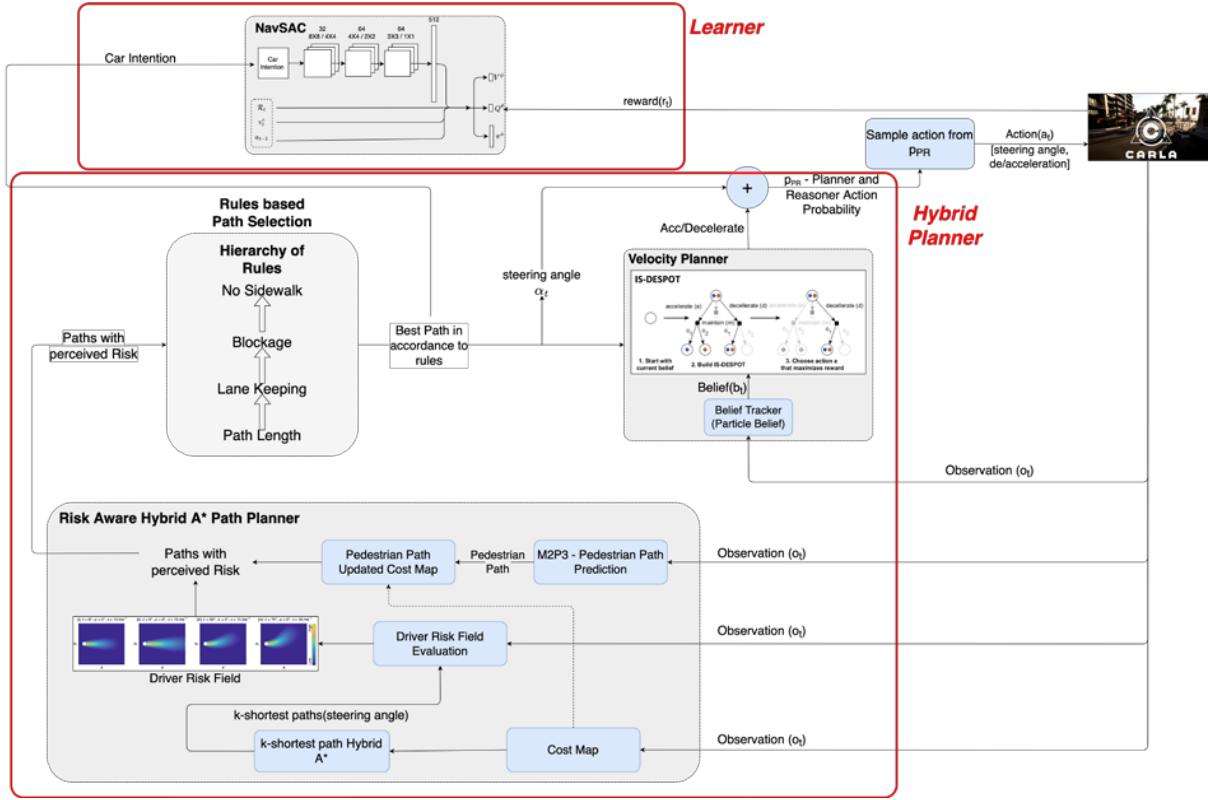

Figure 1: HyLEAR training architecture.

**Training.** The training of HyLEAR with the architecture shown in Figure 1 follows the general interposed learning framework [21]: At each time step $t$, a full car control action $a_t$ is generated by combining steering angle from the selected path and speed action sampled from categorical distribution initialized with the hybrid planner speed policy. Action $a_t$ gets executed in the CARLA driving simulation environment and the respective POMDP belief state transition tuple $(o_t, a_t, R_{t+1}, o_{t+1})$ is stored in the DRL memory buffer $D$. The deep reinforcement learner NavSAC takes as input (a) a small RGB segmentation image as snippet from the planning cost-map that includes observed environmental information of the actual scene together with past and future path of the car and pedestrian, as well as (b) the current reward, speed, and previous action to learn to output the optimal speed action policy. For this purpose, it processes the image with a convolutional neural network, and samples a set of transitions from $D$ for its off-policy soft actor-critic learning with the respective loss functions

$$J_V(\psi) = \mathbb{E}_{o_t \sim D}\left[\tfrac{1}{2}(V^\psi(o_t) - \mathbb{E}_{a_t \sim \pi^\phi}[Q^\theta(o_t, a_t) - \log \pi^\phi(a_t|o_t)])^2\right]$$
$$J_Q(\theta) = \mathbb{E}_{(o_t, a_t) \sim D}\left[\tfrac{1}{2}(Q^\theta(o_t, a_t) - \hat{Q}(o_t, a_t))^2\right];$$
$$J_\pi(\phi) = \mathbb{E}_{o_t \sim D}\left[\log \pi^\phi(a_t|o_t) - Q^\theta(o_t, a_t)\right]$$

where observation $o_t$ and action $a_t$ are sampled from $D$, $V^\psi$ is state value function, $Q^\theta$ is Q-function, $\pi^\phi$ (speed) actor policy function, and $\psi$, $\theta$ and $\phi$ are the NavSAC network weights to be trained. This hybrid planning-assisted informed state-action space sampling may reduce the training time of NavSAC by avoiding unnecessary explorations.

**Testing.** The testing architecture of HyLEAR is the same as for training except that it does not include the computationally expensive planner IS-DESPOT to determine an optimal speed actions for given situation and path. This capability is now embedded in and performed much faster by the trained learner NavSAC. During testing, the hybrid planner of HyLEAR generates the shortest safe path with minimal human-perceived risk and rule violations as input for the trained NavSAC such that the extracted steering angle together with the optimal speed action determined now by the trained NavSAC instead of the planner IS-DESPOT is then executed as a full control action by the car in the driving simulator CARLA.

## 3  Evaluation

**Experimental setting.** Our experimental comparative performance evaluation of HyLEAR has been conducted over the synthetic benchmark CARLA-CTS1, which consists of twelve parameterized scenarios with about thirty thousand scenes in total simulated in the driving simulator CARLA. Most of the traffic scenarios are taken from the GIDAS accident study [2] where the car is confronted with street crossing pedestrians, possibly occluded by some parking car, an incoming car and intersections. The scenes per scenario are generated with varying speed and crossing distance of pedestrians from the car. The selected baselines are (a) the individual CFN action planning and learning methods IS-DESPOT-p and NavSAC-p each of which guided by a hybrid A* path planner, as well as the socially-aware DRL method A2C-CADRL [8] and the hybrid learning-assisted planning method HyLEAP [17] for CFN. The performance of each method is measured in terms of (a) overall safety index (SI) defined as total number of scenarios in which the method is below given percentages of crashes and near-misses (5, 10); (b) crash and near-miss rates, and time to goal (TTG); (c) ride comfort defined as being inversely proportional to the equally weighted sum of jerks and human-perceived risk of planned trajectory, risk normalized to [0,1] with risk threshold set to 0.1 [11]; and (d) training and execution time. HyLEAR is implemented in Python and PyTorch framework; all methods were trained and tested on the DL supercomputer NVIDIA DGX-1 at DFKI Kaiserslautern.

**Results.** The overall results of our experiments, averaged across all scenarios, are shown in Table 1.

| Method | Safety (SI) | Crash (%) | Near-miss (%) | TTG (s) | Comfort | Training (d) | Exec (ms) |
|---|---|---|---|---|---|---|---|
| NavSAC-p | 1 | 21.44 | 9.24 | 17.57 | 0.262 | 10 | 60.29 |
| IS-DESPOT-p | 1 | 21.01 | 6.05 | 16.20 | 0.689 | N/A | 259.56 |
| A2C-CADRL-p | 0 | 25.14 | 11.47 | **14.26** | 1.010 | 4 | **58.57** |
| HyLEAP | 4 | **19.26** | **7.41** | 16.16 | 0.803 | 5 | 215.80 |
| HyLEAR | **5** | 19.88 | 8.49 | 15.86 | **1.064** | **3** | 71.50 |

Table 1: Overview of results on the CARLA-CTS1 benchmark

In general, HyLEAR provided a safer and more comfortable ride than the other selected baselines, following comparatively the fastest training and with a relatively acceptable time to goal and execution time. In some cases, taking the safe and more comfortable but not shortest route by HyLEAR may come at the expense of minimal time to goal compared to the fastest method A2C-CADRL. The latter, however, performed worse on safety due to local minima of always accelerating to reach the goal, which resulted in second best comfort due to zero jerks. On average, the n-step look-ahead action planning with IS-DESPOT-p was as safe as the learner NavSAC-p and with more ride comfort, in particular in scenarios with temporarily occluded pedestrians but required extremely more execution time due to online planning than both DRL methods. The interposed learning allowed HyLEAR to learn the optimal speed action for given situation and safe path faster than the other DRL methods, while due to its hybrid planning HyLEAR performed best in terms of safety and ride comfort, only driving on safe paths through free sidewalks if there are no alternatives with acceptable human-perceived risk. While both hybrid methods, HyLEAP and HyLEAR, are by far more safe than the tested individual planning and DRL methods, the hybrid planning-assisted learning of HyLEAR outperformed the DRL-assisted online planning of HyLEAP in safety, comfort, time to goal, training and execution time.

# 4 Conclusion

We presented HyLEAR, the first hybrid planning-assisted deep reinforcement learning method for collision-free driving policies for self-driving cars that also take into account, whenever possible, the ride comfort and a given set of driving-behavior rules. The experimental results over the CARLA-CTS benchmark revealed that HyLEAR can outperform the selected baselines in safety and ride comfort with faster training and acceptable execution time.

**Acknowledgments.** This work has been funded by the German Ministry for Education and Research (BMB+F) in the project MOMENTUM.